\documentclass[journal]{IEEEtran}
\usepackage[utf8]{inputenc}
\usepackage{graphicx}

\title{Exploring Relations in Untrimmed Videos for Self-Supervised Learning}
\author{Dezhao Luo, Bo Fang, Yu Zhou, Yucan Zhou, Dayan Wu, Weiping Wang
\IEEEcompsocitemizethanks{\IEEEcompsocthanksitem
This work is supported by the National Key R\&D Program of China (2017YFB1002400), the Open Foundation of the State Key Laboratory of Media Convergence and Communication at Communication University of China, the Beijing Municipal Science \& Technology Commission (Z191100007119002), the CCF-Tencent Open Fund, the Youth Innovation Promotion Association CAS, the Excellent Talent Introduction of Institute of Information Engineering of CAS (No. Y7Z0111107), and the Key Research Program of Frontier Sciences, CAS, Grant NO ZDBS-LY-7024.

D. Luo, B. Fang, Y. Zhou, Y. Zhou, D. Wu and W. Wang are with Institute of Information Engineering, Chinese Academy of Sciences, Beijing, 100089, China (e-mail: luodezhao@iie.ac.cn, fangboiie@gmail.com, zhouyu@iie.ac.cn, zhouyucan@iie.ac.cn, wudayan@iie.ac.cn, wangweiping@iie.ac.cn).

D. Luo and B. Fang share the same contribution.

Y. Zhou is the corresponding author.

This work has been submitted to the IEEE for possible publication. Copyright may be transferred without notice, after which this version may no longer be accessible.
}
}
\date{August 2020}

\begin{document}

\maketitle

\begin{abstract}
  Existing video self-supervised learning methods mainly rely on trimmed videos for model training. However, trimmed datasets are manually annotated from untrimmed videos. In this sense, these methods are not really self-supervised.  In this paper, we propose a novel self-supervised method, referred to as Exploring Relations in Untrimmed Videos (ERUV), which can be straightforwardly applied to untrimmed videos (real unlabeled) to learn spatio-temporal features. ERUV first generates single-shot videos by shot change detection. Then a designed sampling strategy is used to model relations for video clips. The strategy is saved as our self-supervision signals. Finally, the network learns representations by predicting the category of relations between the video clips. ERUV is able to compare the differences and similarities of videos, which is also an essential procedure for action and video related tasks. We validate our learned models with action recognition and video retrieval tasks with three kinds of 3D CNNs. Experimental results show that ERUV is able to learn richer representations and it outperforms state-of-the-art self-supervised methods with significant margins.
\end{abstract}

\section{Introduction}
%要解决的问题
%CNNS在计算机视觉领域取得了极大的成功。但是全监督在用cnns做视觉任务的时候，通常在labbeld大数据集上如，imagenet和kinetics上做预训练，来提升cnns的特征提取能力。however，提取这些数据的label非常耗费人力和时间，而且，当出现新领域新任务的时候，以前的label用不上的时候，又要打上新的label。而且，有的复杂数据集，复杂任务的标注是非常困难的。

Convolutional Neural Networks(CNNs) have achieved great success in the computer vision field, especially for image related tasks \cite{krizhevsky2012imagenet}. In general, CNNs are trained with large-scale labeled image datasets such as ImageNet \cite{russakovsky2015imagenet}. If transferred to some downstream tasks ($e.g.,$ object detection, instance segmentation), the pre-trained models can promote the performance since they have better feature extraction capabilities. However, manual annotations of large-scale datasets are time-consuming and expensive, particularly for video related tasks.
%On the contrary, large amounts of unannotated video data can be freely obtained. Using the unannotated video data to train CNN models is of great significance.

%Furthermore, annotating new datasets could always be an obstacle for researches on new domains. Meanwhile, large-scale datasets are potentially to make great steps for video understanding. and Kinetics \cite{kay2017kinetics}%

%现有的方法是怎么做的
%为了解决以上的问题，自监督获得了成功（关注），他通过从数据本身中提取label的方式，来促进网络对unlabeled data的学习。对于图片而言，。。。。。 对于视频而言。。。。
%To overcome the above insufficient of using labeled data and take good advantages of large amounts of videos from website, self-supervised representation learning has achieved unprecedented attention in recent years. They\cite{larsson2017colorization} extract labels from the raw data by  pre-processing, hide the information to generate input data, and then encourage the network to learn the hidden part of input data. In this manner, models are trained to learn representations without human annotations. The trained models are then used to promote downstream tasks. 
\begin{figure}[!t]
\centering
\includegraphics[width=1.1\columnwidth]{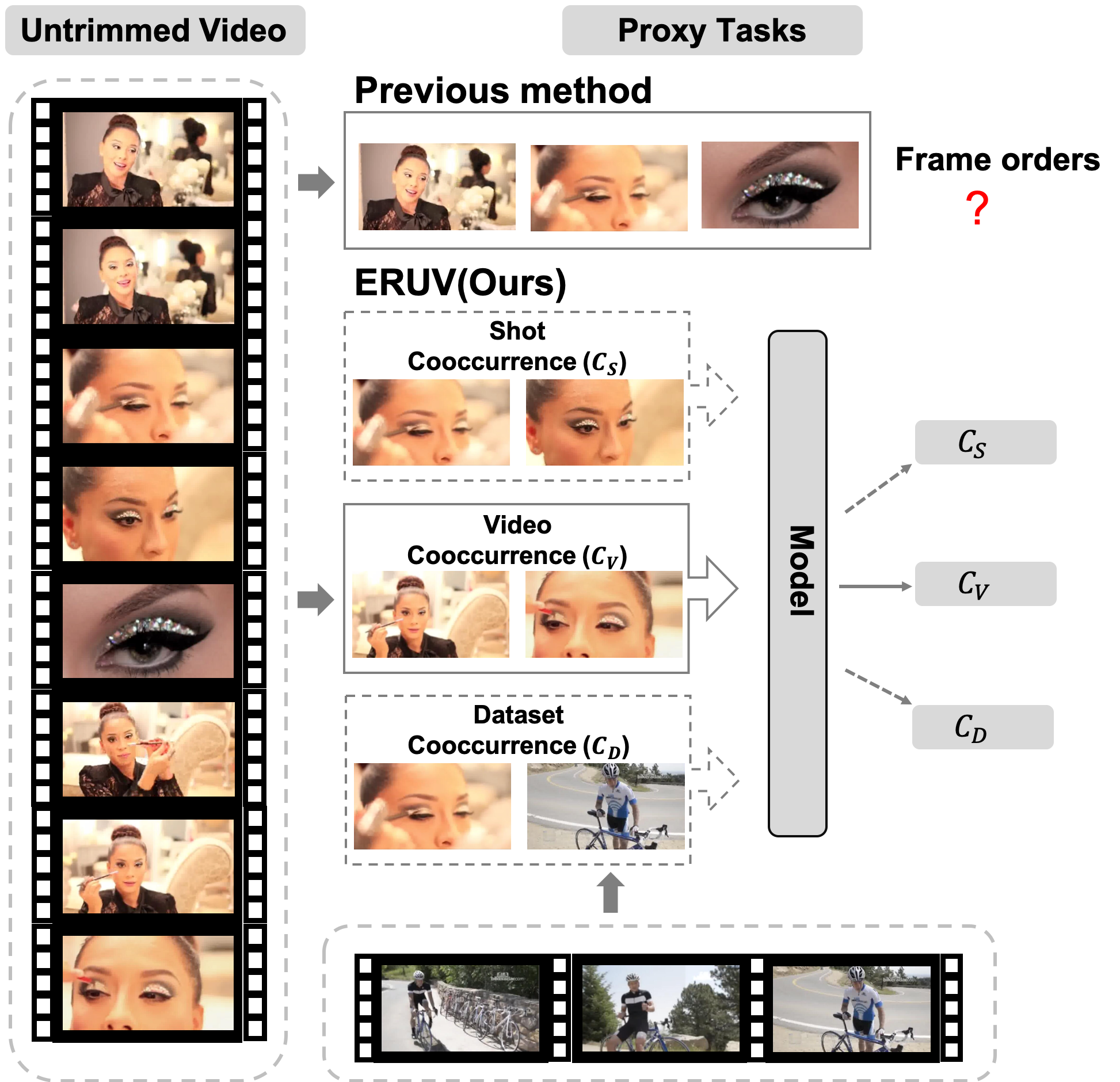}
\caption{Illustration of the necessity to design proxy tasks for untrimmed videos. With shot changes in untrimmed videos, the models of previous methods may be confused while ERUV can take advantage of them.}
\label{fig:First}
%\vspace{-1em}

\end{figure}

Self-supervised representation learning, which extracts the supervisory signal from raw unlabeled data automatically as the learning target, has attracted unprecedented attention in recent years. First, labels are generated from the raw unlabeled data by pre-processing. Then, the labels together with the raw data are input to the network for model training. In this manner, models are trained to learn representations without human annotations. The trained models are then used to promote downstream tasks.

In previous studies on self-supervised learning for images, relative location \cite{doersch2015unsupervised,noroozi2016unsupervised} or color of image \cite{larsson2017colorization} are used as supervision signals. Recently, many approaches have been proposed for video self-supervised learning. \cite{misra2016shuffle,lee2017unsupervised,fernando2017self,wei2018learning,gan2018geometry,buchler2018improving} aim to learn representations for 2D CNNs. However, state-of-the-art performance for video related tasks are mostly based on 3D CNNs \cite{tran2015learning,tran2018closer,feichtenhofer2019slowfast}. For better representing the spatio-temporal dynamics of videos with 3D CNNs in self-supervised manner, 3D video cubic puzzles \cite{kim2019self}, video motion and appearance statistics \cite{wang2019self}, video clip orders \cite{xu2019self} and video cloze procedure \cite{luo2020video} are taken as the supervisory signals.

%the orders of frames are used for videos \cite{kim2019self,fernando2017self,misra2016shuffle} . , Mas \cite{wang2019self} is proposed to predict motion and appearance statistics of videos. To focus on moving proportion of the visual field, models have to learn spatio-temporal features that can identify the foreground and background.
%现有方法的缺陷
%现存的视频自监督方法，大部分都是用的trimmed video来做。他们忽略了，trimmed video也是人为的从untrimmed video  中剪辑而来的，并抛弃了视频中的与运动无关的clip.it may be difficult to adapt lar in more reaalsize and hallenfing secaoes 。1是youtube上面的视频是untrimmed的by nature，二是 有的复杂行为的边界不好界定。

The existing video self-supervised methods learn spatio-temporal representations with trimmed video datasets (e.g., UCF101 \cite{soomro2012ucf101}, HMDB51 \cite{jhuang2011large}, and Kinetics \cite{kay2017kinetics}). Nevertheless, trimmed video datasets are not real unlabeled because the start and the end frames of action instances are annotated manually. As is shown in Fig.\ref{fig:First}, genuine untrimmed videos may include action foreground frames and background frames simultaneously, and may include multi-view camera shots for the same action instance. As a result, previous 3D CNNs self-supervised learning methods are invalid with untrimmed videos. Inspired by exploring the relations of objects in object recognition \cite{felzenszwalb2009object,hu2018relation}, we argue that the relations between actions within different shots in untrimmed videos may supply informational supervisory signals for self-supervised representation learning.

%They ignored that trimmed videos are also artificially convert from untrimmed video. The videos in those datasets have been trimmed to regions with activities occurring, and the background of the videos are discarded during the course of annotation, which are not real unlabeled data.  Although \cite{piergiovanni2019evolving} pre-trains with random Youtube videos, it designs confused tasks for untrimmed video. We can't simply apply previous methods to untrimmed videos, since some of the background frames are sport-independent and different actions in untrimmed videos are order-independent , Fig.\ \ref{fig:First}.  

%我们的方法是什么
%为了解决这个问题，我们提出了ERUV。在ERUV中，我们从untrimmed video中采样一组视频，并且训练一个网络比较这些视频的区别和联系，从而学习到丰富的特征。这种比较机制在这个论文中已经讨论过了，我们定义了丰富的关系比较，可以促进网络学习到更多的特征 

In this paper, we propose a novel self-supervised representation learning approach referred  to as Exploring Relations in Untrimmed Videos (ERUV), targeting at learning representations while comparing differences and similarities between video clips. In ERUV, we generate video clips with a designed sampling strategy to model different relations. Then, we train a 3D-CNNs model to identify the categories of the relations. This mechanism has been explored by \cite{hu2018relation} in image object detection, while we extend it with a self-supervised manner to model the relations between video clips. Moreover, modeling rich and complicated relations of videos can promote the network's spatio-temporal representation capability.

%understanding of videos.

%我们怎么解决的
%具体的，ERUV利用了untrimmed video中的包含镜头转换的特点，实现了不同的采样和组合策略，以定义不同的关系。我们包含三个部分，relation building，feature extrction，comparing。 第一个部分。。。第二个部分。。。第三个部分。。。
%关系
Specifically, ERUV consists of three components including shot editing, relation modeling and video comparing. The first component generates single-shot videos with shot change detection, since the single-shot videos focus on consistent motions. The second component promotes exploring representation capability by modeling cooccurrence and relevance relations between sampled video clips. Finally, video comparing model learns representation by predicting the categories of relations.

%区分
The contributions of this work include:

\begin{itemize}
%\vspace{-1em}

\item We propose ERUV to capture video appearance and temporal representations. To the best of our knowledge, this is the first self-supervised representation learning work utilizing untrimmed video datasets, so ERUV is a real self-supervised video representation learning method. 
%我们通过给clip构建关系的方式，学习视频的内容一致性
\item We propose a novel feature learning strategy. By modeling relations between video clips, we can integrate current self-supervised methods with our designed relations.
\item Extensive experimental results demonstrate that the trained networks learn rich spatio-temporal representations, and the proposed method outperforms other recently proposed self-supervised learning methods considerably.
\end{itemize}

\begin{figure*}[!t]
     \centering
     \includegraphics[width=2.0\columnwidth]{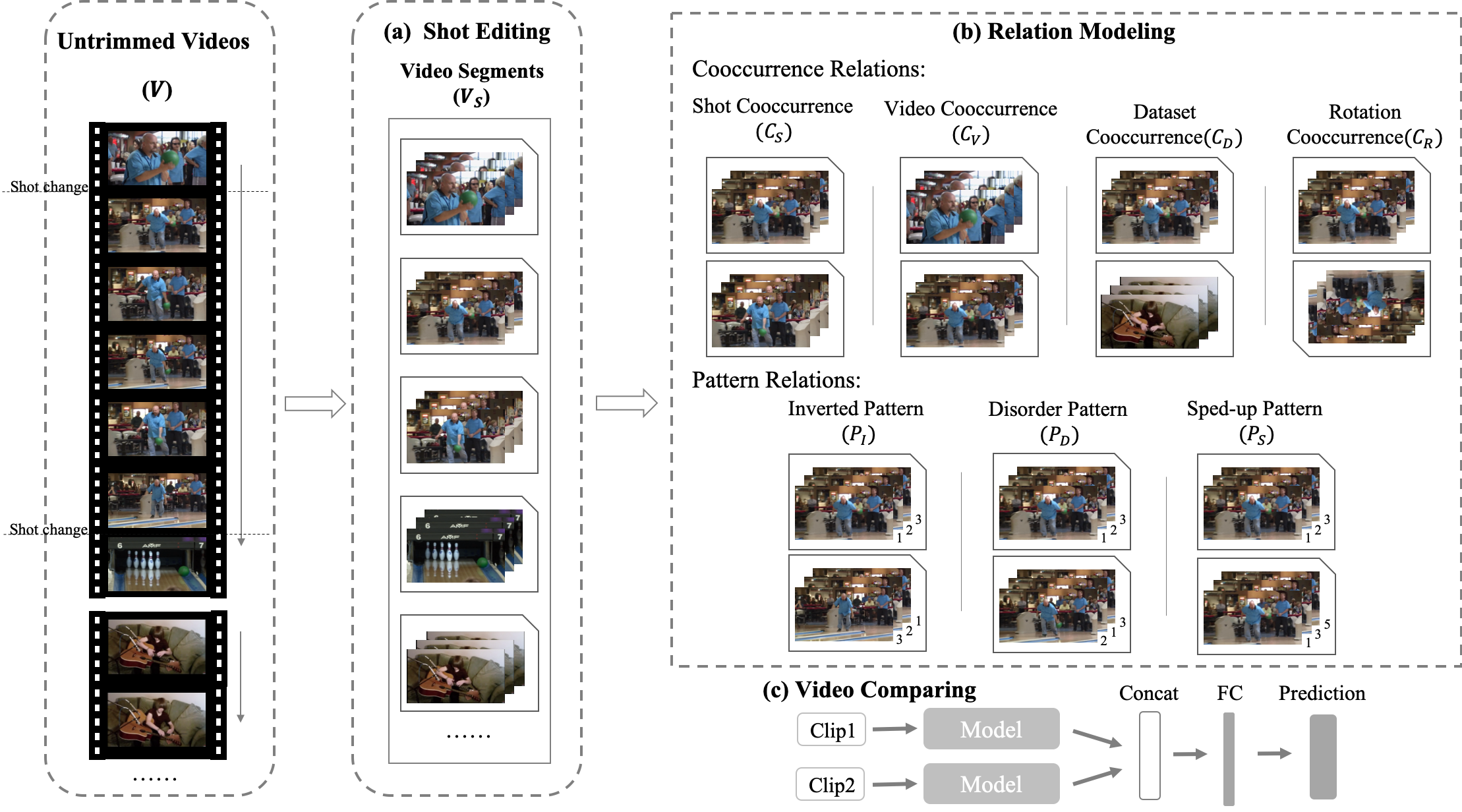}
     \caption{Illustration of the ERUV framework. Given untrimmed videos $V$, ERUV first generates video segments ($V_S$) by shot change detection and long-duration shot breakdown. Then the relation modeling is used to sample video segments with different relations. The sampling strategy is stored to be the label. Finally the clips extracted from the video segments are fed to 3D models for spatio-temporal representation learning.}
     \label{fig:framework}
 \end{figure*}

\section{Related Work}
%引出下面要讲的三个部分
%Self-supervised is well studied in the domain of video understanding. Since it  extracts labels from data itself and obtains corresponding knowledge without human annotations, it mainly used to boost the task of action classification tasks. In this section, we will review the most recent and relevant papers related to our approach.%
The most relevant works to ours are video action recognition and self-supervised representation learning, which are introduced in the following two subsections respectively.
\subsection{Video Action Recognition}
%介绍行为识别
Great progress has been achieved in video action recognition with deep neural networks. 

Earlier researches take videos as sequences of frames, and apply 2D CNNs
to extract features. 
\cite{simonyan2014two} proposes two-stream convolutional networks in which RGB frames and optical flows are processed by 2D CNNs. 
Temporal segment network \cite{wang2016temporal} is based on a sparse temporal sampling strategy and enables efficient and effective learning with the whole action video level supervision. 
In \cite{zhou2018temporal}, temporal relation reasoning is proposed to learn and reason the temporal dependencies among video frames of different time scales.
%generates optical flow with RGB frames and take it as another input stream with RGB frames. These two streams are individually processed by the same 2D CNNs which are then fused to predict the video category.%

Recently, methods based on 3D CNNs have become mainstream because they can model both spatial and temporal features simultaneously. C3D \cite{tran2015learning} extends the 2D convolution kernels to 3D kernels to model temporal features among frames. \cite{tran2018closer} proposes R3D and R(2+1)D. R3D is an extension of ResNet to 3D. In R(2+1)D, 3D convolution kernels are decomposed for spatial convolution and temporal convolution. Slowfast \cite{feichtenhofer2019slowfast} uses a slow pathway with low frame rate to model spatial semantics and a fast pathway with high frame rate to capture temporal motion information. 

%All these methods perform well when trained with annotated trimmed video data, and self-supervised representation learning with untrimmed videos is still an open problem.

%Targeting to model video features both in spatial and temporal, \textcolor[rgb]{1,0,0}{C3D,I3D, R3D, R(2+1)D \cite{tran2015learning,tran2018closer}} are proposed.%

%3d
\subsection{Self-Supervised Representation Learning}
Self-supervised learning extract information from unlabelled data to train models. Previous methods usually encourage models to learn rich  representations by predicting information which is hidden within un-annotated data. Afterwards, the learned models can be used to promote the performance of downstream tasks.  Recently, some self- supervised learning methods with the supervision signal obtained automatically from unlabeled images or videos have attracted much attention. 
\subsubsection{Image Representation Learning}
In order to produce supervision signals for images, spatial transforms are usually applied to pre-process the unlabeled images \cite{larsson2017colorization,lee2019multi,pathak2016context}.  For instance, the related position between image patches are applied as the signals. \cite{noroozi2016unsupervised,doersch2015unsupervised} leverage image information by predicting relative positions of image patches. \cite{Inpainting2016} take the color of images as label, it encourage the network to learn statistic features via image colorization task.
%While \cite{pathak2016context} trains models to decode the content of a withheld image region according to its surroundings.

\subsubsection{Video Representation Learning}
Generally, in self-supervised video representation learning, the supervisory signal is generated automatically from unlabeled videos without manual intervention.

%helps to release the insufficient of labeling data for fully supervised learning both in image and video domains. Over the last few years, several self-supervised tasks has gain much attention. Unlabeled images are applied with spatial pre-processing to generates supervision signals. \cite{noroozi2016unsupervised,doersch2015unsupervised} leverage image information by predicting relative positions of image patches. \cite{Inpainting2016} take the color of images as label, it encourage the network to learn stastic features via image colorization task. While \cite{pathak2016context} trains models to decode the content of a withheld image region according to its surroundings.%

Early approaches mostly focus on self-supervised learning of 2D CNNs. \cite{misra2016shuffle,lee2017unsupervised} utilizes the orders of frames as the supervision signals. \cite{fernando2017self} proposes an odd-one-out network to predict the unrelated clip over a set of video clips. 
\cite{wei2018learning} exploits the arrow of time as a supervisory signal. 
%for the activity analysis. 
\cite{gan2018geometry} extracts pixel-wise geometry information as flow fields and disparity maps and uses them as auxiliary supervision. %\cite{buchler2018improving}, a reinforcement learning method is used to generate more effective permutations for the 2D CNNs self-supervised learning.

Recently, several methods for 3D CNNs self-supervised learning are proposed to learn the complicated spatial and temporal representation \cite{zhao2017spatio,vondrick2016generating}. \cite{kim2019self} proposes a video representation learning method for 3D CNNs based on solving 3D video cubic puzzles. \cite{jing2018self} proposes 3DRotNet in which the supervisory signals are rotation angles. \cite{wang2019self} proposes to learn 3D CNNs representations by predicting the motion and appearance statistics of  unlabeled videos. In \cite{xu2019self}, the order of video clips is used as the supervisory signal for 3D CNNs' training.  \cite{luo2020video} proposed to complete a video cloze procedure for representation learning.
%To predict the motion region and the appearance colors, the network has to learn spatio-temporal temporal features which can identify the foreground and background.% 

Despite the effectiveness of existing methods, they are all based on trimmed video datasets such as UCF101 \cite{soomro2012ucf101}, which are not really unlabeled.
%In fact, to get the trimmed videos, the start and end frames of the action must be annotated manually. 
However, videos are genuinely untrimmed in real applications, in which previous self-supervised learning methods are invalid, as shown in Fig.\ref{fig:First}. Therefore, we argue that it is of great significance for developing a self-supervised learning method to train 3D CNNs with untrimmed video datasets.

%unfortunately ignore that  the datasets they use for self-supervised training are human trimmed, which are not real unlabeled data. Moreover, their proposed task can't be simply applied to untrimmed videos. Therefore, we suggest that it is critical for developing a self-supervised method to train 3D CNNs with real unlabeled datasets.
%

\section{Exploring Relations in Untrimmed Videos}
%描述方法 relation已经被正式有效，我们将其转移到视频领域来。介绍nlp的attention，aggeration model 一起被学,讲motivation
Modeling relations for objects would help object recognition \cite{felzenszwalb2009object}. Specifically, it predicts how likely two object classes may appear in the same image. The process of modeling relations for objects is comparing the differences and similarities between them, which can boost the feature extraction capability of the network.
%第一段，简要描述方法的三个部分
%在这一章节中，我们先简单介绍一下整体的模型，然后将从细节上介绍每个方法部分。如图2所示，采样用的是。。。第二个用的是，第三个用的是。。。。
Motivated by the success of modeling relations for objects, we propose a novel representation learning method, referred to as ERUV.

ERUV consists of three components: shot editing, relation modeling, and video comparing, which is shown in Fig.\ref{fig:framework}. In shot editing, the  untrimmed videos are cut into single-shot videos based on shot change detection, then the long-duration shot is cut into several video segments. Whether the video segments are from the same untrimmed video or the same video shot are stored for generating the supervisory signals. For relation modeling, targeting to generate video clips with different relations, clips are sampled from video segments with designed sampling strategies. In video comparing, we use 3D CNNs to extract spatio-temporal representations for the sampled clips, then the extracted features are concatenated and fed to a fully-connected layer to predict the possible relation categories.

\subsection{Shot Editing}
\label{sec:greetings} 
%和trimmed video 不同的是，trimmed video包含了及其复杂的运动变化，不仅有多个运动instance，还有许多与运动无关的视频。因为复杂，所以要裁剪
%而要确保内部一致，要基于镜头来裁剪，尽头裁剪方式如下。
 Different from trimmed videos, an untrimmed video often exhibits extremely complex dynamics. It may consist of both action foreground frames and action background frames, and the action foreground frames may only occupy small portions of the whole video sequence. However, the network needs to learn spatio-temporal features of continuous motion patterns, in which no shot change should exist. In order to generate video clips focusing on a continuous motion pattern, we first edit videos with shot-based processing.  
 
 Given an untrimmed video $V$ from datasets $\mathcal{D}$, we take $V$ as a sequence of frames $V=\{f_t\}, t \in \{ 1,2,...,\mathcal{T} \}$, where $\mathcal{T}$ is the total number of frames in $V$. HOG features for ${f_t}$ are firstly extracted and then HOG feature difference is calculated between each adjacent frame $f_t$ and $f_{t+1}$ as a metric to measure the change of frame appearance. If the absolute value of HOG difference is larger than a given threshold, we take it as a shot change between $f_i$ and $f_{i+1}$. After shot change detection, to breakdown long-duration actions, the shots are further cut into short video segments with a fixed length of $K$. These $K$-frame video segments are denoted as  $V_{S}$. Suppose we have a shot denoted as $S^j=(b_j,e_j)$, where $(b_j,e_j)$ represents the beginning and ending location of the $j^{th}$ shot $S^j$, we produce video segments from this shot as $V_{S}(S^j)=\{(b_j+i \times K,b_j+(i+1) \times K)\}$, where $i \geq 0$ and $b_j+(i+1) \times K \leq e_j$. In the end, all these video segments from different shots are merged, $V_S$ can be denoted as $V_S = \{ v_n \},  0 \le n \leq \mathcal{N} $, where $\mathcal{N}$ is the total number of segment videos in $V_S$. We allocate the shot id and the video id to each video segment to mark the source shot and the source untrimmed video it is extracted from.
 
 Thus, we have obtained single-shot video segments with continuous motion patterns without human annotation. To be noted, we only clip the untrimmed videos with automatic shot change detection. Without discarding any background frames or breaking down any actions by their temporal borders, $V_{S}$ is different from human trimmed videos.

%为了训练网络来比较不同的clip，从而识别他们之间的关系。我们设计来co-currence 和区别度关系。为了学习到深度的特征，选项应该是有效的迷惑学习者，同时又保留了视频特征。在这基础之下，我们设计了。我们设计了6种关系，其中有不同的情况来迷惑。

\subsection{Relation Modeling}
In order to train the network to compare differences and similarities between video clips to learn spatio-temporal features, we design  spatial cooccurrence and temporal pattern relations for self-supervised learning, which correspond to the learning of spatial  and temporal features. The relations designed for self-supervised learning should be simple yet effective so that the model is able to learn rich representations. As shown in Fig.\ref{fig:framework}, We design 7 kinds of relations: shot cooccurrence ($C_D$), video cooccurrence ($C_V$), dataset cooccurrence ($C_D$) and rotation cooccurrence ($C_R$)  for spatial learning; Inverted pattern ($P_I$), disorder 
pattern ($P_D$), sped-up pattern ($P_S$) for temporal learning.

ERUV takes 2 video segments as input and the predicted relation category as output. The video segments are sampled from $V_{S}$, denoted as $v_i$ and $v_j$, where $i,j \in \{n\}$. The relation between $v_i$ and $v_j$ are stored as our labels. In our implementation, $v_i$ is randomly selected from $V_S$ for each training step, $v_j$ is sampled according to its relation with $v_i$. The relation details and the corresponding sampling strategies of $v_j$ are described as follows.
%To be noted, to make a fair prediction, the number of each relation samples should be the same during the training step.

%为了提供一种能够使模型关注到表观的特征，我们提出了co-occureence.我们通过判断，这个模型是否出现在一个视频中，是否出现在不同视频中，还是同一个运动instance中。是否出现在一个视频中，需要很高的语义信息。
\subsubsection{Spatial Cooccurrence Relation} To provide relations that focus on spatial representation learning, we introduce cooccurrence relations.
%共存性关注视频的表观特征，并评估两段视频是来自哪一种共存性关系。
ERUV can make the prediction of a specific cooccurrence relation by measuring spatial similarities between the inputs. In order to generate videos of different apparent similarities, we  propose shot cooccurrence $C_S$, video cooccurrence $C_V$ , dataset cooccurrence $C_D$ and rotation cooccurrence $C_R$.
%how likely two video segments can occurrence in a same untrimmed video or a same video shot. 

$C_S$ denotes that the actions in $v_i$ and $v_j$ may occur in the same  shot, which means their spatial features are almost the same. Given a video segment $v_i$ during the training step, ERUV randomly samples $v_j$ from the video segment sets which have been labeled to be extracted in the same shot as $v_i$.

$C_V$ denotes that the actions in $v_i$ and $v_j$ occur in the same video but in different shots, they may describe the same action which is taken by different cameras from different angles or two relative actions in semantics. 
%For $C_V$, $v_i$ and $v_j$ are generated from different  shots, but these shots come from the same untrimmed video.  

For $C_D$, $v_i$ and $v_j$ are generated from  different untrimmed videos, which denotes the actions between them will not occur in the same untrimmed video.

For $C_R$, $v_i$ is randomly rotated by 90, 180, or 270 degrees to generate $v_j$, so that the model is forced to learn orientation related features. 

 It is worthwhile to note that predicting cooccurrence relations between video segments $v_i$ and $v_j$ requires high-level semantic appearance information, and understanding the structure of objects or colors is not enough to tackle this task.
%为了提供关注到。。时许。。我们提供了这种选项
\subsubsection{Temporal Pattern Relation}
To provide relations that focus on temporal features, we further introduce three kinds of temporal pattern relations, in which the video clips have similar appearances but different temporal patterns of actions, including invert pattern $P_I$, disorder pattern $P_D$ and sped-up pattern $P_S$. 
Given a video segment sampled from $V_S$, denoted as $v_i$, we apply a temporal transformation on it to generate $v_j$. The corresponding transformation to each relation is described as follows.

For $P_I$, $v_j$ is a temporally inverted version of $v_i$. For $P_D$, it denotes that $v_j$ is a disordered version of $v_i$. To generate $v_j$, we shuffle the frames of $v_i$ randomly. For sped-up pattern $P_S$, $v_j$ is a fast-forward version of $v_i$. We adapt uniform sampling with an interval of $s$ on $v_i$, which is denoted as $s \times$ dilated sampling. The procedure generates $v_j$ with $s \times$ fast-forward playback rate. In our implementation, $s$ could be 2 or 4. Fig.\ref{fig:framework} shows an example of $s=2$. To distinguish the minor difference between temporal patterns, the network has to learn the temporal representations. 

\begin{figure}[!t]
\centering
\includegraphics[width=1.0\columnwidth]{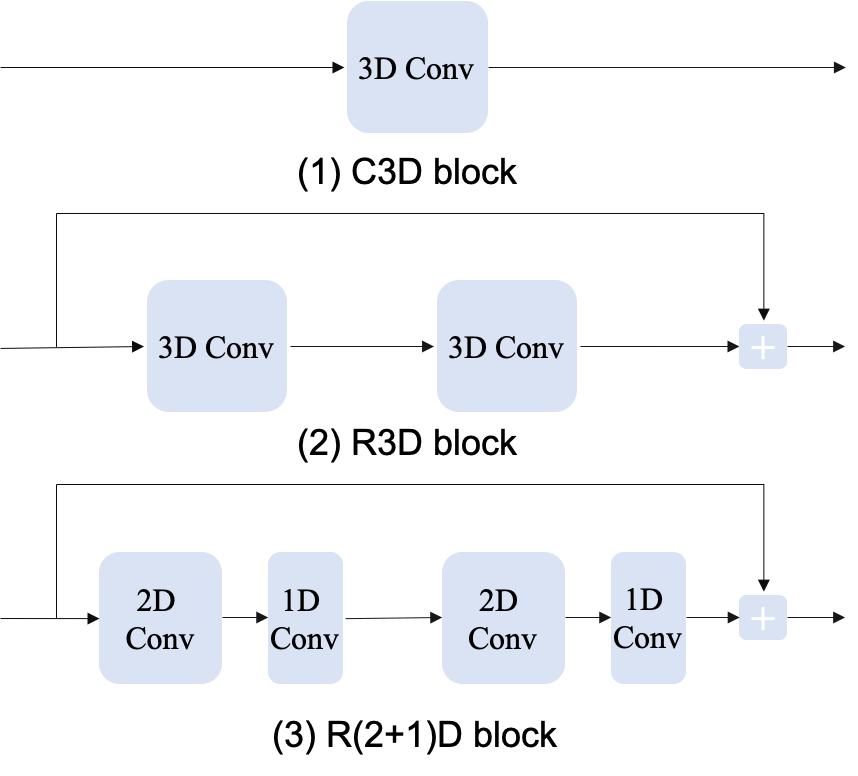}
\caption{Illustration of C3D, R3D, and R(2+1)D blocks.
}
\label{fig:model}
\vspace{-1em}

\end{figure}

\subsection{Video Comparing}
Given $v_i$, $v_j$ and their relation by the previous section, ERUV randomly generates a $k$-frame clip from each video segment as the learning sample to fed to backbones. 

To learn a feature representation from video comparing, we take it as a classification task and use a simple siamese network. This network has 2 parallel stacks of layers with shared parameters. Every network stack takes C3D \cite{tran2015learning}, R3D or R(2+1)D \cite{tran2018closer} as the backbones to extract both spatial and temporal features. Each stack takes a video clip as input and produces a representation as output.  

The structure of the backbones is shown in Fig.\ref{fig:model}. Since 2D CNNs are able to obtain spatial information, C3D extents 2D CNNs for spatio-temporal representation learning as it can model the temporal information of videos. It stacks five C3D blocks which consist of a classic 3D convolution. R3D is an extension of C3D, which refers to 3D CNNs with residual connections. To be specific, R3D block consists of two C3D blocks,  the input and the output are connected by a residual unit. Besides, in R(2+1)D, the difference between the R(2+1)D block and the R3D block is that the 3D convolution is decomposed into a spatial 2D convolution and a temporal 1D convolution.

The features extracted from 3D backbones are then concatenated  to a linear classification layer. 
 The output is a probability distribution over different relations. With $a_i$ is the $i$-$th$ output of the fully connected layer for relations, the probabilities are as follows:
$$p_i=\frac{\exp(a_i)}{\sum^{c}_{j=1}\exp(aj)}$$

\noindent where $p_i$ is the probability that the relation belongs to class $i$, and $c$ is the number of relations. We update the parameters of the network by minimizing the regularized cross-entropy loss of the predictions:
 $$\mathcal{L} = -\sum^c_{i=1}y_i\log(pi)$$

\noindent where $y_i$ is the groundtruth. 

While this network uses 2 clips at training time, during testing we can obtain 3D CNNs representations of a single clip by using just one stack because the parameters across the stacks are shared.

%这一部分描述我们到底学到了什么，以及背景帧的相关问题
%1,解释背景帧不是噪音，加入背景帧学习的必要性
%背景帧中有场景和运动物体。虽然运动的物体没有发生具体的运动，但是学习他们的特征能够帮助理解行为信息，我们通过讲背景帧和运动信息建立 video -relation的方式，讲运动视频和背景视频关联起来。，
\begin{figure}[!t]
\centering
\includegraphics[width=1.0\columnwidth]{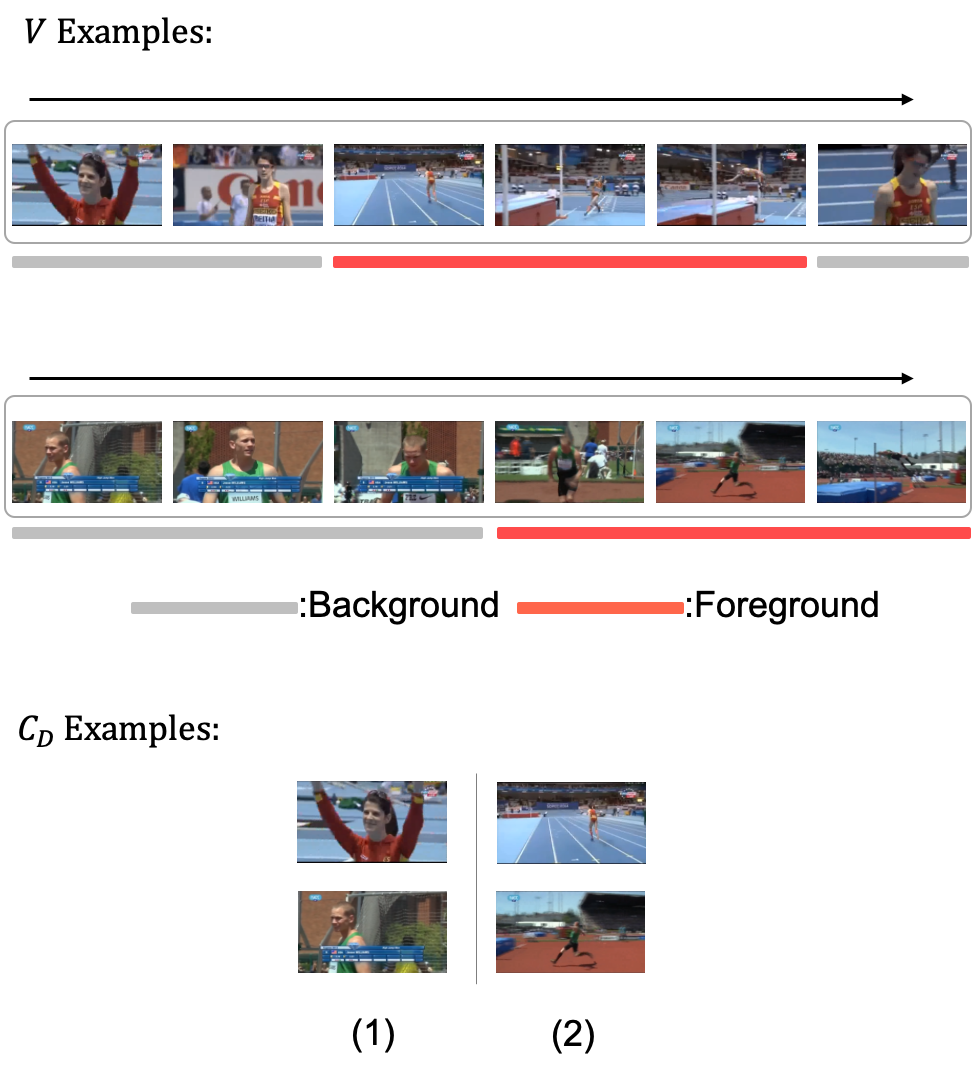}
\caption{Illustration of $C_D$ examples from high jump. (1) shows an example of  all the videos are background, (2) are foreground. ERUV is able to learn even if all videos are from the same category no matter they are background or foreground.
}
\label{fig:example}
\end{figure}
\subsection{Discussion}
%我们提出一种新颖的自监督方式学习特征。与之前方法【vcp】不同的是，在空间上，我们提出了对空间要求更高的cooccurence relation。在时间上，提出了sped-up pattern。
%在我们的训练过程中，背景帧也是宝贵的训练样本。如图所示，就算两个视频都是北京，那我们也可以学到是否来自同一个镜头。需要注意的是，我们的标签定义的是，是否来自同一个视频，并没有定义是否来自同一个行为。如果不同的视频中，出现了相似的背景，我们也鼓励网络去学习该数据集这样的一个分布特点。
ERUV contributes to the domain of video self-supervised learning in two aspects.
Firstly, compared with previous methods \cite{xu2019self,luo2020video}, we propose novel and effective relations on videos. The cooccurrence relations will encourage the models not only to understand the internal spatial features but also to learn discriminative features by predicting their spatial relations which require rich spatial semantic features.  
%1，我们的目标是要学习untrimmed video，这些video里面很多背景帧
%2，就算遇到背景帧我们也可以学习
%3，如图，都是背景且相同行为也可以根据运动目标或者背景判断
%4，如果都是背景，通过pattern learning 也能学到

Also, ERUV targets in learning features with untrimmed videos which contain many background clips. As shown in Fig.\ref{fig:example}, if all videos are from the same category, we can also classify it to $C_D$ by their moving objects or scenes. Similarly, the pattern relations can still be helpful when the videos are backgrounds because the purpose of pattern relations is to gain the capacity to generate temporal features, rather than learning some specific action patterns.

%2，解释我们为什么能够做到区别背景帧的关系
%背景帧只要有运动轨迹，我们就能从relation中区分出来，而eruv的共存性，学习的是数据集中的视频分布，学习到的是是否来自同一个视频，而不是学习是否来自同一个类，我们讲每一个视频看作一个类。
%4, 学到了什么
%通过时间的关系学到类运动特征，通过空加你关系，学到类表冠特征。
%目的
%怎么做的
%介绍concatnate
%\subsection{Discussion}
%自监督要学习有辨识度的特征，从时许和空间上，有能区分不同的视频的区别和联系。
 %Downstream tasks, $e.g.,$ video retrieval, action recognition are essentially video classification tasks. Previous methods design tasks for understanding spatio-temporal features within video clips. Differently, ERUV proposes to learn discriminative  features by comparing videos, which conforms to the procedure for video classification.

\section{Experiment}
In this section, we demonstrate the effectiveness of ERUV. First, We elaborate the experimental settings. Second, we conduct ablation studies to quantify the contributions.
Third, we compare the representations of the network with other approaches and visualize them for clarity. Finally, we treat our method as a self-supervised approach to initialize models for action recognition and video retrieval and compare it with state-of-the-art methods.

\subsection{Experimental Settings}
\subsubsection{Datasets}
We pre-train our method on one large-scale dataset, namely Thumos14 \cite{jiang2014thumos}. The dataset is suitable for our method as they provide the original untrimmed videos.
The Thumos14 dataset has 101 classes for action recognition and 20 classes for action detection. It is composed of four parts: training data, validation data, testing data, and background data. To verify the effectiveness of our method, we use the validation data (1010 videos) to train ERUV.

The experiments are fine-tuned on UCF101 and HMDB51 datasets to evaluate the performance of our self-supervised pre-trained network. UCF101 consists of 101 action categories with about 9.5K videos for training and 3.5k videos for testing. It exhibits challenging problems including intra-class variance of actions, complex camera motions, and cluttered backgrounds. HMDB51 consists of 51 action categories with about 3.4k videos for training and 1.4k for testing.The videos are mainly collected from movies and websites including the Prelinger archive, YouTube, and Google videos.

\subsubsection{Network Architecture}
 For video representation extraction, we choose C3D, R3D, and R(2+1)D as backbones in ERUV.
 C3D stacks five 3D convolution blocks, each block consists of a classic 3D convolution with the kernel size of $3\times 3\times 3$ and followed by a batch normalization layer and a ReLU layer. R3D block consists of two 3D convolution layers followed by batch normalization and ReLU layers. The input and output are connected with a residual unit. R(2+1)D decompose the 3D kernel to a spatial $1\times 3\times 3$ kernel and a temporal  $3\times 1\times 1$ kernel.

\subsubsection{Implementation Details}
  The validation set of Thumos14 is used to train our self-supervised learning method, while UCF101 and HMDB51 are used to validate the effectiveness of ERUV.
  
  In video editing, we follow the settings in \cite{wang2017untrimmednets} for shot change detection. And long-duration shots are broken down with a fixed length of $K= $ 300 to generate $V_{S}$. Those videos in $V_{S}$ which are shorter than 48 frames are discarded. The length $k$ of each clip is set to be 16, corresponding to the inputs of most 3D CNNs. Each frame is resized to $128\times171$ and randomly cropped to $112\times112$. We set the initial learning rate to be 0.01, momentum to be 0.9. Our pre-training process stops after 300 epochs and the best validation accuracy model is used for downstream tasks.

\begin{table}
    \caption{Accuracy of relation classification. ``$C_S$'' denotes the shot coocurrence, ``$C_V$'' the video coocurrence, ``$C_D$'' the dataset coocurrence, ``$C_R$'' the rotation coocurrence,``$P_I$'' the inverted pattern, ``$P_D$'' the disorder pattern, ``$P_S$'' sped-up pattern.}
    \resizebox{1.0\columnwidth}{!}{
    \centering
    
    \begin{tabular}{lcccccccc}
    \hline

     Method&  Overall(\%)& $C_S$(\%) & $C_V$(\%) & $C_D$(\%) & $C_R$(\%)& $P_I$(\%) & $P_D$(\%) & $P_S$(\%) \\
    \hline
    C3D & 66.7& 71.2& 89.3&84.5 & 76.2  &50.5&44.3&48.5\\
%    \hline
%    R(2+1)D &78.45& 60.43 &87.81& 99.61& 48.18&96.33 \\
    \hline
    \end{tabular}
    }

    \label{fig:eruv acc}
    \vspace{-1em}

\end{table}
 \begin{table}[!t]
    \centering
    \caption{Ablation study of cooccurrence and pattern relations. ERUV are firstly pre-trained on Thumos14 and then used to fine-tune action recognition on UCF101. The figures refer to action recognition accuracy.}
    \begin{tabular}{lc}
        \hline
    Method  &UCF101 acc(\%)  \\
            \hline

        Random initialization          & 62.1 \\
        \hline
        ERUV with $C_{S,D}$  & 65.0 \\
        ERUV with $C_{S,R}$ & 66.3 \\
        ERUV with $C_{S,V}$ & 67.9 \\
        ERUV with $C_{S,V,D,R}$  & 68.8 \\
        \hline
        ERUV with $P_{D,S}$ & 66.0 \\
        ERUV with $P_{I,S}$& 65.0 \\
        ERUV with $P_{I,D}$  & 65.4 \\
        ERUV with $P_{I,D,S}$    & 66.4 \\
        \hline
        ERUV with $C_{S,V,D,R}P_{I,D,S}$&
        \textbf{70.4} \\
        \hline

        \end{tabular}

    \label{table:relations}
    \vspace{-1em}

\end{table}

\subsection{Ablation Study}
In this section, we evaluate the effect of our designed relations on the first split of the UCF101. We first perform self-supervised pre-training using the validation data in Thumos14. The learned weights are then used as initialization for the supervised action recognition task.

Table \ref{fig:eruv acc} shows the results on Thumos14 which are trained and evaluated on the validation data. It can be seen that ERUV achieves 66.7\% overall accuracy,  for shot coocurrence ($C_S$), video coocurrence ($C_V$), dataset coocurrence ($C_D$) and rotation coocurrence($C_D$), ERUV respectively achieves 71.2\%, 89.3\%, 84.5\%, and 76.2\% accuracy. Considering that the accuracy of random guessing for the task is 14.3\% (7 relations), the framework indeed learns to analyze the content of clips. Besides, it also shows that the designed relations are plausible.

%\subsubsection{Hyper-Parameters}
%We train a model to classify co-occurrence and relevance relations targeting to determine the clip number $N$ and the concatenation strategy. We choose to compare our method between 2-clips and 3-clips, since 4-clips requires a longer video length, which will drastically reduce our data samples. To fix the number of relations for 2-clips and 3-clips, we set $c_j$ to be $c_1$ when generating relevance relations. We evaluate the video comparing task on a held-out validation set. As shown in Table \ref{table:hyper}, we can see from the results that ERUV model can always outperform the random model, which means that ERUV can learn effective spatio-temporal representation.

%As the clip numbers increase, the relation classification accuracy
%increases from 33.1\% (2-clips simple) to 45.9\% (3-clips simple) and the action accuracy obtains an increment of 1.0\% (from 63.9\% to 64.9 \%), which denotes that the relations between more clips can promote the network representation. We also compare the feature pairwise concatenation strategy with the simple concatenation strategy. With the pairwise strategy, the relation classification accuracy increase from 45.9\% to 56.7\% while the action accuracy increase from 64.9\% to 66.7\%. According to these results, 3 clips and pairwise concatenation are used in the following experiments.  

As shown in Table \ref{table:relations}, to clearly show the effect of relations for representation learning, we conduct ablation experiments on ERUV with various relations for action recognition. 

It can be seen that when pre-training with $C_{S,D}$ ($C_S$ and $C_D$ only), $C_{S,R}$ or $C_{S,V}$, the accuracy of action recognition outperforms the baseline (Random initialization) by 2.9\%, 4.2\% or 5.8\%. When using all cooccurrence relations (ERUV with $C_{S,V,D,R}$), the performance further increased to 68.8\%.  Pre-training with  $P_{D,S}$ ($P_D$ and $P_S$ only) improves the performance by 3.9\%, When using all pattern relations (ERUV with $P_{I,D,S}$), the performance further increased to 66.4\%. Combining the cooccurrence relation and pattern relation (ERUV with  $C_{S,V,D,R}P_{I,D,S}$) finally improves the performance to 70.4\%, significantly outperforming the baseline by 8.3\%. The experiments show that ERUV can learn representative features and hence to promote the performance on action recognition task. 
 \begin{figure}[!t]
\centering
\includegraphics[width=1.0\columnwidth]{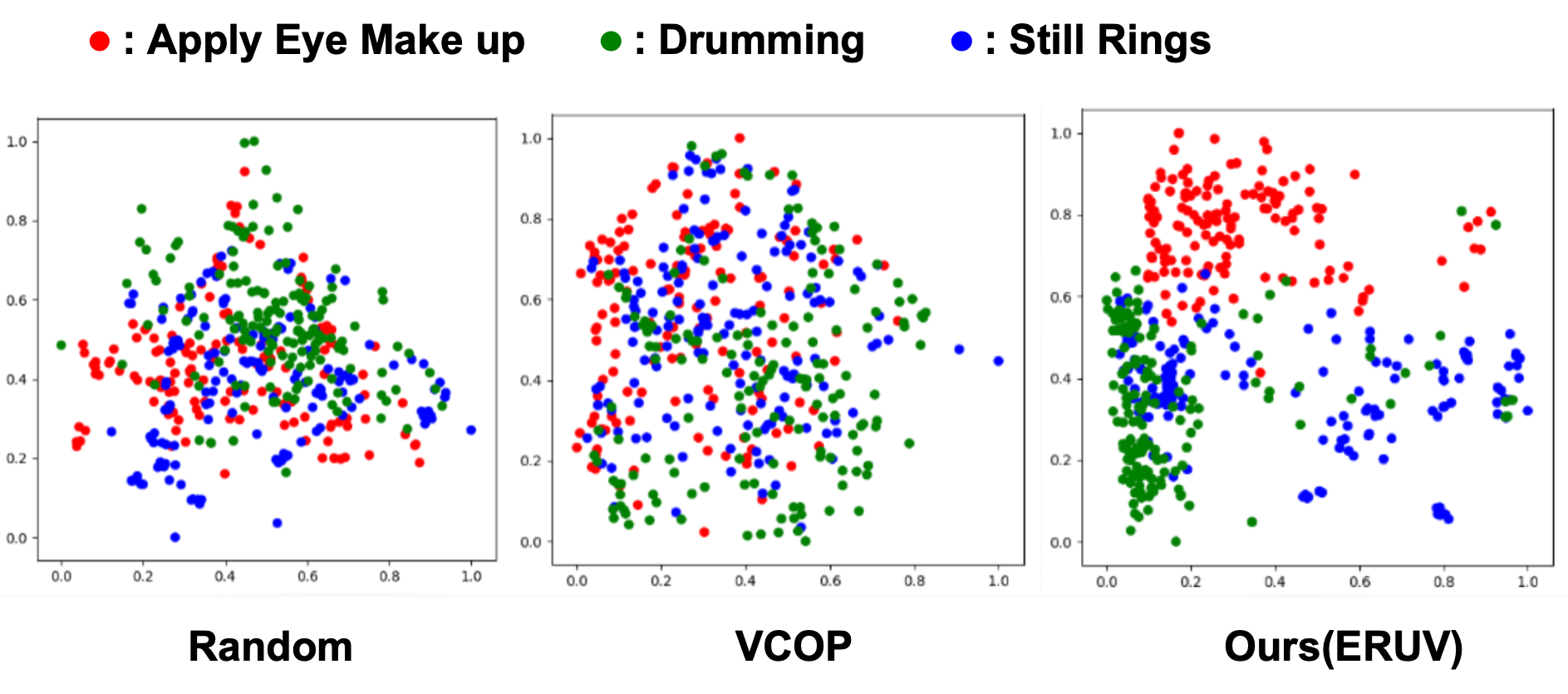}
\caption{Feature embedding results with ERUV compared with the random method and SOTA method VCOP.} 

\label{fig:4}
\vspace{-1em}
\end{figure}

 \begin{table}[!t]
    \centering
    \caption{Performance comparison under different pre-training datasets. UCF101(Thumos14) denotes the model is pre-trained on Thumos14 and fine-tuned on UCF101.}
    \begin{tabular}{ccc}
        \hline
    Method  &UCF101(UCF101)(\%)&UCF101(Thumos14)(\%)  \\
            \hline

    VCOP \cite{xu2019self}&     66.7  & 64.9\\
        VCP \cite{luo2020video} & 69.7 &68.7 \\
                \hline

        ERUV& \textbf{71.2} &\textbf{70.4}\\

        \hline

        \end{tabular}

    \label{table:retrain}
    \vspace{-1em}

\end{table}

\subsection{Representation Learning}
In this section, we further demonstrate the effectiveness ERUV comparing with previous methods on different training strategies.

First of all, as shown in Table \ref{table:retrain}, VCOP \cite{xu2019self}, VCP \cite{luo2020video} and ERUV are trained with UCF101 and Thumos14 respectively. Then the trained models are used to initialize the action recognition task on the first split of UCF101. In our implementation, VCOP and VCP are trained with $V_S$, which are generated by shot editing \ref{sec:greetings}. Since it seems impossible for them to learn with raw untrimmed videos.  Besides, when trying to train ERUV with UCF101, we generate $C_S$ in the same video, also we remove the relation of $C_V$ and $C_D$, because there is no such label in the dataset. 
    
It can be seen that when pre-training and fine-tuning on UCF101, ERUV outperforms VCOP by 4.5\% and VCP by 1.5\%. Showing that relations used in ERUV are better than previous methods. Pre-training on Thumos14 and fine-tuning on UCF101, ERUV can also outperform VCOP and VCP by 5.5\% and 1.7\% respectively. Note that, because we choose to fine-tune on UCF101, all methods perform better when pre-training on UCF101 than on Thumos14. 

%The relations in ERUV is targeted at learning discrimintative features. 
To indicate that why ERUV can gain better performance on action recognition, we visualize the features generated by the trained models. To be specific,  we select 300 samples of 3 action classes from UCF101 and visualize their pool5 features with two-dimensional embeddings by PCA. 

In Fig.\ref{fig:4}, we can see that the random method can not extract effective discriminative features. For VCOP \cite{xu2019self}, the inter-instance distance increases but it still can not extract effective discriminative features. For ERUV, the intra-class distance increases, and the inter-instance distance decreases. This implies that our model can learn better discriminative features, which can directly promote the action recognition task.

\subsection{Target Tasks}
%The above experiments show ERUV can learn rich and useful representation. 
To further validate the effectiveness of ERUV, we use the pre-trained model to initialize action recognition backbones and directly apply the extracted features to video retrieval.  
\subsubsection{Action Recognition}
\begin{table}
    \centering
     \caption{Comparison of action recognition accuracy on UCF101 and HMDB51. 
     %VCOP* is implemented according to the official code.
     }
        \resizebox{1.0\columnwidth}{!}{

    \begin{tabular}{lcc}
    \hline
    Method&UCF101(\%)&HMDB51(\%)  \\
    \hline
    Jigsaw \cite{noroozi2016unsupervised}&51.5&22.5 \\

    OPN \cite{lee2017unsupervised}&56.3&22.1\\
    B\"uchler\cite{buchler2018improving}  &58.6&25.0 \\
     Mas\cite{wang2019self} & 58.8&32.6\\
    3D ST-puzzle\cite{kim2019self}& 65.0 &31.3\\
    \hline
    C3D(random) &61.8&24.7\\
    C3D(VCOP \cite{xu2019self}) &65.6& 28.4\\ 
    C3D (VCP) \cite{luo2020video} &68.5&32.5\\
    C3D(ERUV) &\textbf{69.6}&\textbf{33.7}\\ 
    \hline
    R3D(random) &54.5&23.4\\
    R3D(VCOP \cite{xu2019self}) &64.9& 29.5\\ 
    R3D(VCP \cite{luo2020video}) &66.0& 31.5\\ 
    R3D(ERUV) &\textbf{68.8}& \textbf{31.6}\\ 
    
    \hline
    R(2+1)D(random) &55.8&22.0\\
    R(2+1)D(VCOP \cite{xu2019self}) &\textbf{72.4}& 30.9\\ 
   % R(2+1)D(VCOP*) &61.0& 28.8\\ 
    R(2+1)D(VCP \cite{luo2020video}) &66.3& \textbf{32.2}\\ 

    R(2+1)D(ERUV) &68.4& 31.9\\ 
    \hline
    \end{tabular}}

    \label{fig:state of the art}
    \vspace{-1em}
\end{table}

After training with untrimmed videos from Thumos14, we fine-tune the model using labeled videos. We implement the fine-tune procedure and follow the settings of \cite{xu2019self}. 
%During fine-tune we initialize the backbones from ERUV, and randomly initialize the fully-connected layers.
The training step for fine-tune stops after 150 epochs. During the test, we sample 10 clips uniformly for each video to obtain the action prediction.

Table \ref{fig:state of the art} shows the results on the UCF101 and HMDB51, we report the averages accuracy over 3 splits.
It can be seen that, with C3D backbones, ERUV obtains 69.6\% accuracy compared with 61.8\% of random initialization on UCF101 dataset, 33.7\% to 24.7\% on HMDB51. 
It also outperforms the state-of-the-art VCP approach \cite{luo2020video} by 1.1\% and 1.2\% respectively.
ERUV also achieves better accuracy with R3D and R(2+1)D backbones. With R3D$\backslash$R(2+1)D backbones, ERUV has 14.3\%$\backslash$12.6\% and 8.2\%$\backslash$9.9\%  performance gains over random method on UCF101 and HMDB51 respectively.

Since our training samples of Thumos14 maintains a large portion of background videos, the improvements gained on 3 splits of UCF101 and HMBD51 datasets show the effectiveness and generalization of ERUV. 

\subsubsection{Video Retrieval}
To directly test the features extracted by ERUV, we validate the pre-trained model with nearest-neighbor video retrieval.
Following the protocol in \cite{xu2019self}, we extract 10 clips for every video with the ERUV pre-trained model. The clips extracted from the test set are used to query the $k$-$th$ nearest clips from the training sets.
 If a video of the same category is matched, a correct retrieval is counted.

The video retrieval results on UCF101 and HMDB51 are shown in Table \ref{fig:retrieve ucf101} and \ref{fig:retrieval hmdb}, which further indicate the effectiveness of ERUV trained models. Note that we outperform the SOTA method dramatically with different backbones on top1(\%), for which the features extraction ability is critical.

\begin{table}
    \caption{Video retrieval performance on UCF101.}

    \resizebox{1.0\columnwidth}{!}{
    \centering

    \begin{tabular}{lccccc}
    \hline
    Methods&top1(\%)&top5(\%)&top10(\%)&top20(\%)&top50(\%)  \\
    
        \hline
    Jigsaw\cite{noroozi2016unsupervised} &19.7&28.5&33.5&40.0&49.4 \\
    OPN\cite{lee2017unsupervised} &19.9&28.7&34.0&40.6&51.6 \\
    B\"uchler\cite{buchler2018improving} &25.7&36.2&42.2&49.2&59.5 \\
        \hline
    C3D(random) &16.7&27.5&33.7&41.4&53.0 \\
    
    C3D(VCOP\cite{xu2019self}) & 12.5 & 29.0&39.0&50.6&66.9\\
    C3D(VCP\cite{luo2020video}) & 17.3 & 31.5&42.0&52.6&67.7\\

    C3D(ERUV)&\textbf{25.2}&\textbf{40.5}&\textbf{48.3}&\textbf{57.6}&\textbf{70.4} \\
    \hline
    R3D(random) &9.9&18.9&26.0&35.5&51.9 \\
    R3D(VCOP\cite{xu2019self}) & 14.1 & 30.3&40.4&51.1&66.5\\
    R3D(VCP\cite{luo2020video}) & 18.6 & 33.6&42.5&\textbf{53.5}&68.1\\

    R3D(ERUV)&\textbf{ 21.4}&\textbf{35.2}&\textbf{43.8}&53.1&\textbf{68.3} \\
    \hline
    R(2+1)D(random) &10.6&20.7&27.4&37.4&53.1 \\
   R(2+1)D(VCOP\cite{xu2019self}) & 10.7 & 25.9&35.4&47.3&63.9\\
   R(2+1)D(VCP\cite{luo2020video}) & 19.9 & 33.7&42.0&50.5&64.4\\

     R(2+1)D(ERUV)& \textbf{22.0}&\textbf{35.1}&\textbf{42.6}&\textbf{51.5}&\textbf{64.9} \\
    \hline
    \end{tabular}
    }
    \label{fig:retrieve ucf101}
    \vspace{-1.5em}

\end{table}

\begin{table}
    \caption{Video retrieval performance on HMDB51.}

    \resizebox{1.0\columnwidth}{!}{
    \centering

    \begin{tabular}{lccccc}
    \hline
    Methods&top1(\%)&top5(\%)&top10(\%)&top20(\%)&top50(\%)  \\
        \hline
    C3D(random) &7.4&20.5&31.9&44.5&66.3 \\
    C3D(VCOP\cite{xu2019self}) & 7.4 & 22.6&34.4&48.5&70.1\\
    C3D(VCP\cite{luo2020video}) & 7.8 & 23.8&35.3&49.3&71.6\\
    C3D(ERUV) &\textbf{8.6}&\textbf{25.3}&\textbf{37.0}&\textbf{53.7}&\textbf{75.7}\\

    \hline
    R3D(random) &6.7&18.3&28.3&43.1&67.9 \\
    R3D(VCOP\cite{xu2019self}) & 7.6 & 22.9&34.4&48.8&68.9\\
    R3D(VCP\cite{luo2020video}) & 7.6 & \textbf{24.4}&\textbf{36.3}&\textbf{53.6}&\textbf{76.4}\\

    R3D(ERUV) & \textbf{8.6}&24.3&35.5&53.0&75.9 \\

    \hline
            
    R(2+1)D(random) &4.5&14.8&23.4&38.9&63.0 \\
    R(2+1)D(VCOP\cite{xu2019self}) & 5.7 & 19.5&30.7&45.8&67.0\\
    R(2+1)D(VCP\cite{luo2020video}) & 6.7 & 21.3&32.7&49.2&73.3\\

    R(2+1)D(ERUV)&\textbf{9.3}&\textbf{26.4}&\textbf{38.5}&\textbf{51.8}&\textbf{73.9} \\
    \hline
    \end{tabular}}
    \label{fig:retrieval hmdb}
    \vspace{-1em}

\end{table}

\subsubsection{Visualization}
\begin{figure}[!t]
     \centering
     \includegraphics[width=1.0\columnwidth]{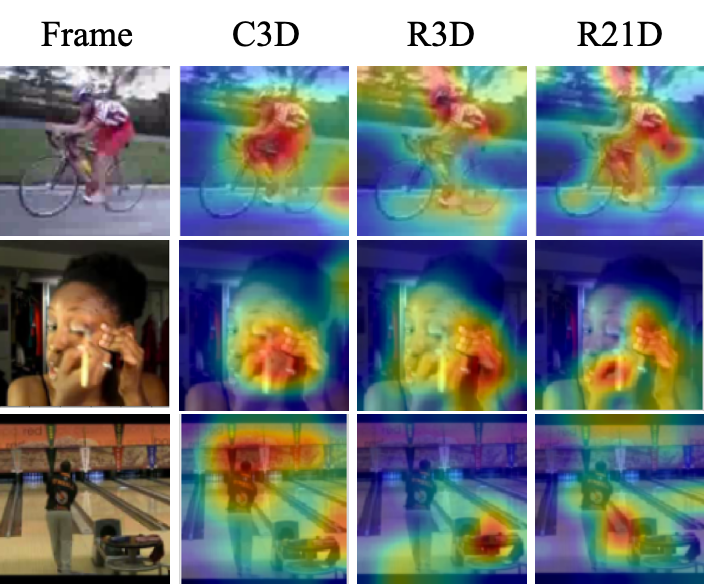}
     \caption{Visualization of attention maps. From left to right: a frame from a video clip, the attention map generated from C3D, R3D and R21D \cite{zagoruyko2016paying}.}
     \label{fig:map}
        \vspace{-1em}

 \end{figure}

    In order to obtain a better understanding of what ERUV learns, we visualize the feature attention maps \cite{zagoruyko2016paying} to indicate where the spatio-temporal representation focuses on. As shown in Fig.\ref{fig:map}, we visualize computed heat maps over sampled frames and compare them under different backbones. It can be seen that the learned features are more likely to focus on the dominant moving objects in the video.

\section{Conclusion}
In this paper, we propose a novel and real self-supervised method referred to as ERUV to obtain rich spatio-temporal features without human annotations. In ERUV, we train CNNs models to predict relations between video clips. Experimental results show the effectiveness of ERUV for downstream tasks such as action recognition and video retrieval. Our network inspires the field of video understanding with two aspects: self-supervised learning can be implemented with untrimmed videos and action relations are beneficial for video understanding.

\bibliographystyle{IEEEtran}
\bibliography{main}

\end{document}